\documentclass[pdflatex,sn-mathphys-num]{sn-jnl}

\usepackage{graphicx}%
\usepackage{multirow}%
\usepackage{amsmath,amssymb,amsfonts}%
\usepackage{amsthm}%
\usepackage{mathrsfs}%
\usepackage[title]{appendix}%
\usepackage{xcolor}%
\usepackage{textcomp}%
\usepackage{manyfoot}%
\usepackage{booktabs}%
\usepackage{algorithm}%
\usepackage{algorithmicx}%
\usepackage{algpseudocode}%
\usepackage{listings}%

\usepackage{todonotes}
\usepackage[justification=centering]{caption}
\usepackage{subfig}

\theoremstyle{thmstyleone}%
\theoremstyle{thmstyletwo}%

\theoremstyle{thmstylethree}%

\newcommand{\orcID}[1]{\href{https://orcid.org/#1}{\includegraphics[width=0.03\textwidth]{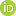}}}

\begin{document}

\title[Vision-based automatic fruit counting with UAV]{Vision-based automatic fruit counting with UAV}


\author*[1]{\fnm{Hubert} \sur{Szolc} \orcID{0000-0003-3018-5731}}\email{szolc@agh.edu.pl}

\author[1]{\fnm{Mateusz} \sur{Wasala} \orcID{0000-0002-8631-8428}}\email{wasala@agh.edu.pl}

\author[1]{\fnm{Remigiusz} \sur{Mietla} \orcID{0009-0001-1893-8748}}\email{mietla@student.agh.edu.pl}

\author[1]{\fnm{Kacper} \sur{Iwicki} \orcID{0009-0000-6550-7476}}\email{kacperiwicki@student.agh.edu.pl}

\author[1]{\fnm{Tomasz} \sur{Kryjak} \orcID{0000-0001-6798-4444}}\email{tomasz.kryjak@agh.edu.pl}

\affil[1]{\orgdiv{Embedded Vision Systems Group, Computer Vision Laboratory, Department of Automatic Control and Robotics}, \orgname{AGH University of Krakow}, \orgaddress{\street{al. Mickiewicza 30}, \city{30--059 Krakow}, \country{Poland}}}



\abstract{The use of unmanned aerial vehicles (UAVs) for smart agriculture is becoming increasingly popular.
This is evidenced by recent scientific works, as well as the various competitions organised on this topic.
Therefore, in this work we present a system for automatic fruit counting using UAVs.
To detect them, our solution uses a vision algorithm that processes streams from an RGB camera and a depth sensor using classical image operations.
Our system also allows the planning and execution of flight trajectories, taking into account the minimisation of flight time and distance covered.
We tested the proposed solution in simulation and obtained an average score of 87.27/100 points from a total of 500 missions.
We also submitted it to the UAV Competition organised as part of the ICUAS 2024 conference, where we achieved an average score of 84.83/100 points, placing 6th in a field of 23 teams and advancing to the finals.}




\maketitle

\section{Introduction}\label{sec:intro}


Unmanned aerial vehicles (UAVs, also known as drones) have recently gained a significant interest from both scientific and commercial communities.
This is due to the relatively wide range of applications, including military missions, cargo transport, terrain inspection, environmental mapping and spectacular cinematography.
One interesting area where UAVs can bring tangible benefits is in smart agriculture \cite{Srivastava2023}, particularly through the use of image processing algorithms \cite{Su2023}.


The potential impact of UAVs in smart agriculture is confirmed, among other things, by recent competitions dedicated to this topic.
One of the most relevant examples is the Unmanned Aerial Vehicle (UAV) Competition accompanying the International Conference on Unmanned Aircraft Systems (ICUAS) 2024.
The challenge was to count different types of fruit using a quadcopter equipped with an RGB camera and a depth sensor.
For this purpose, the organisers prepared a special simulation environment \cite{Markovic2023} to prototype and evaluate solutions.

\begin{figure}
    \centering
    \includegraphics[width=1\linewidth]{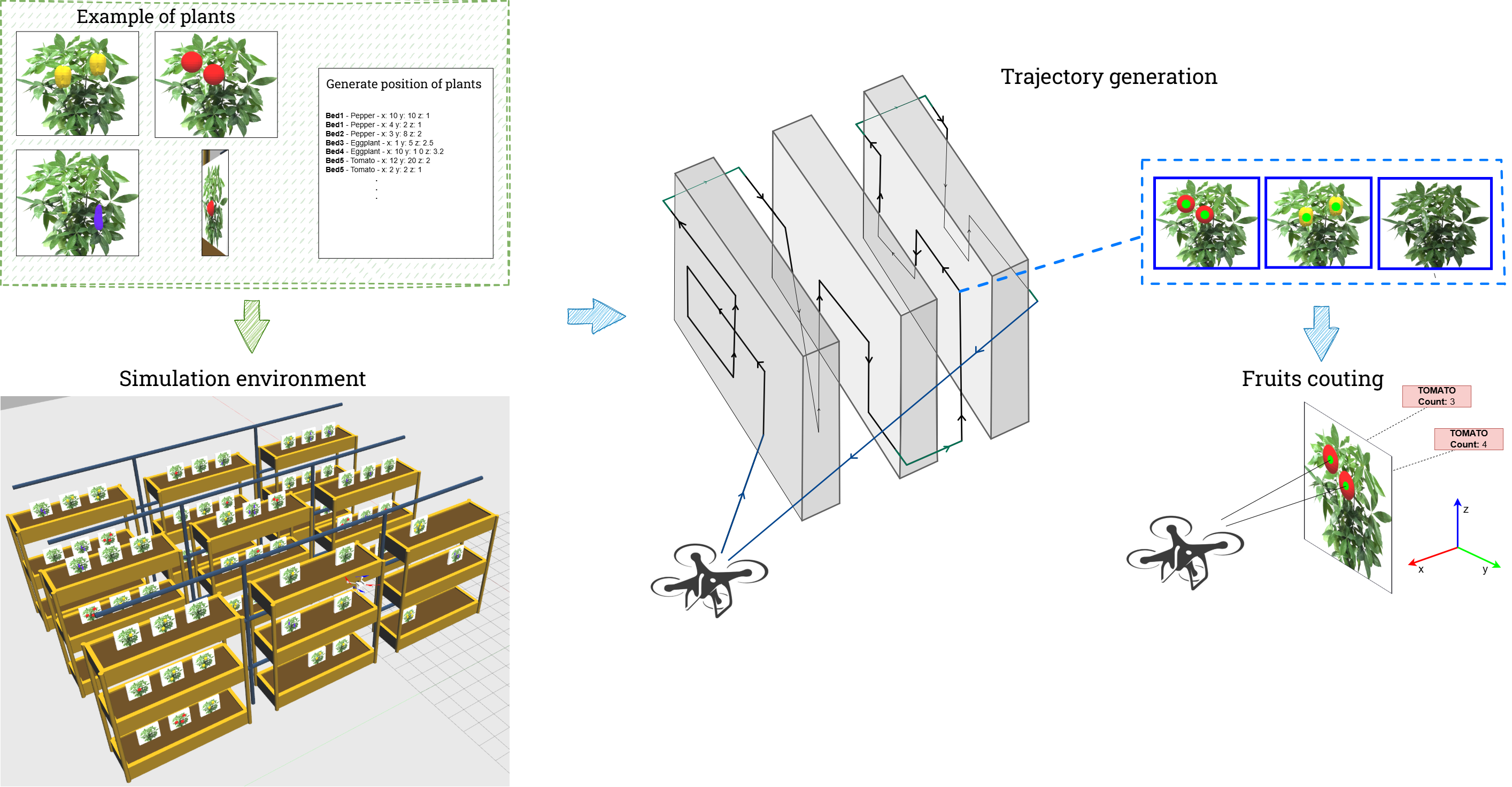}
    \caption{The overview of our vision-based system for automatic fruit counting with UAV. Proposed solution uses a streams from an RGB camera and a depth sensor and process them with classical image processing algorithms. Additionally, our system enables planning and execution of flight trajectories, optimizing for minimal flight time and distance covered.}
    \label{fig:abstract}
\end{figure}

{
Automatic fruit counting is an important task for several reasons.
First of all, it facilitates the new branch of agriculture called \textit{smart agriculture}.
Thanks to fruit counting, it is possible to monitor yields more accurately, to predict the size of the harvest in order to optimise the logistics and sales of the collection, and to detect early fruit drop or uneven ripening.
It also helps to monitor the health of the crop, for example identifying nutrient deficiencies by analysing the distribution and size of the fruit.
For large orchards, fruit counting provides a continuous update on the quantity of fruit, which is impractical to achieve manually.
An automated system to perform this task is therefore essential, especially in large-scale agriculture.
}


In this paper, we present our solution to the aforementioned competition task.
For fruit detection, we propose a novel vision system that uses classical video stream processing algorithms.
Thus, it provides a relatively short processing time to count objects while the UAV is flying.
For route planning, we use the Fast-TSP algorithm \cite{fast-tsp}.
It ensures minimisation of flight time and distance covered, which is associated with reduced energy consumption.
The fruit counting system that we have prepared is shown schematically in Figure \ref{fig:abstract}.
Our solution enabled us to reach the 6th place (out of 23 participants) in the classification of the simulation phase of the competition.

The main contribution of this paper can be summarised as follows:
\begin{itemize}
    \item We have developed {a proprietary fruit detection algorithm which, together with the control strategy, forms a} system for automatic fruit counting using an unmanned aerial vehicle.
    \item We have tested our approach in a simulation environment according to the rules of UAV Competition at ICUAS 2024.
    
\end{itemize}
We make our entire code available under an open licence in the Github repository\footnote{Removed for blind revision}.

The remainder of this paper is organised as follows.
In Section~\ref{sec:related work} we describe similar approaches to automatic object counting with the use of unmanned aerial vehicles.
Our own solution to this problem is presented in Section~\ref{sec:solution}.
In Section~\ref{sec:results} we present results obtained with our method along with a discussion.
The paper concludes with a summary in Section~\ref{sec:summary}, in which we also outline plans for further work.

\section{Related work}\label{sec:related work}


The use of unmanned vehicles in smart agriculture is growing rapidly, offering new opportunities for crop monitoring, plant health assessment and yield counting. 
These methods are gaining popularity due to their efficiency, high accuracy and ability to perform large-scale analysis at relatively low cost.


The paper \cite{fruitnerf2024} presents FruitNeRF, an innovative fruit counting system using Neural Radiance Fields (NeRF).
It enables fruit detection and classification based on different images of the same plant.
The system combines advanced image segmentation (e.g. SAM and U-Net) with the NeRF method to create 3D point clouds representing the fruit.
In addition, the counting process uses cascade clustering, which accurately identifies individual fruits and their clusters based on spatial volume. 
Tests were carried out on synthetic data (e.g. apples, plums, mangoes) and real data (apple trees in an orchard), achieving high accuracy (F1 score: up to 0.95 for reference masks). 
In both cases, the input data was collected by manual camera positioning without the use of drones.


The paper \cite{monocular} proposes a fruit counting system using only a single camera mounted on an unmanned ground vehicle (UGV).
A Faster-RCNN network is used to detect fruits and tree trunks. 
They are then tracked in subsequent images using a Kalman filter and optical flow estimation.
A key part of the solution is the transformation of object tracks from the 2D plane to 3D landmarks.
This eliminates the problem of double fruit counting.
The solution, tested on mango fruit data, achieved results comparable to more sophisticated systems that also use other sensors.


The paper \cite{dronearm} presents a comprehensive Drone-Bee system for autonomous fruit picking in environments without access to a GNSS signal, such as vertical farms. 
The authors developed a lightweight grapple (to be mounted on a UAV), a precision localisation system using SLAM, vision-based perception and a drone flight control algorithm.
The system was optimised to take into account the limited computational power of on-board devices and enables precise fruit picking in challenging conditions such as external disturbances (e.g. wind) or restricted workspaces. 
The Drone-Bee uses the proprietary Fast Fruit Detector (FFD) algorithm \cite{ffd}, which is based on the Detection Transformer (DETR) model and was implemented on the Nvidia Jetson NX platform.


The cited work shows that there is currently considerable interest in the subject of object counting for smart agriculture.
This is particularly true for the fruit detection and localisation algorithms, which are dominated by approaches based on image processing algorithms.
In one of the cited papers, the authors presented a complete counting system using a UAV for this purpose.
However, there is no commonly used dataset, so it is difficult to directly compare different methods.

\section{Proposed solution}\label{sec:solution}


The solution we developed meets the requirements of the UAV competition organised as part of the ICUAS 2024 conference.
We use classical image processing operations.
This ensures short processing times and enables image analysis during the flight.
It also reduces the power consumption of the processing unit.
We also take into account the need to plan the flight path of the UAV in order to check all the set locations.


\begin{figure}
    \centering
    \includegraphics[width=0.8\linewidth]{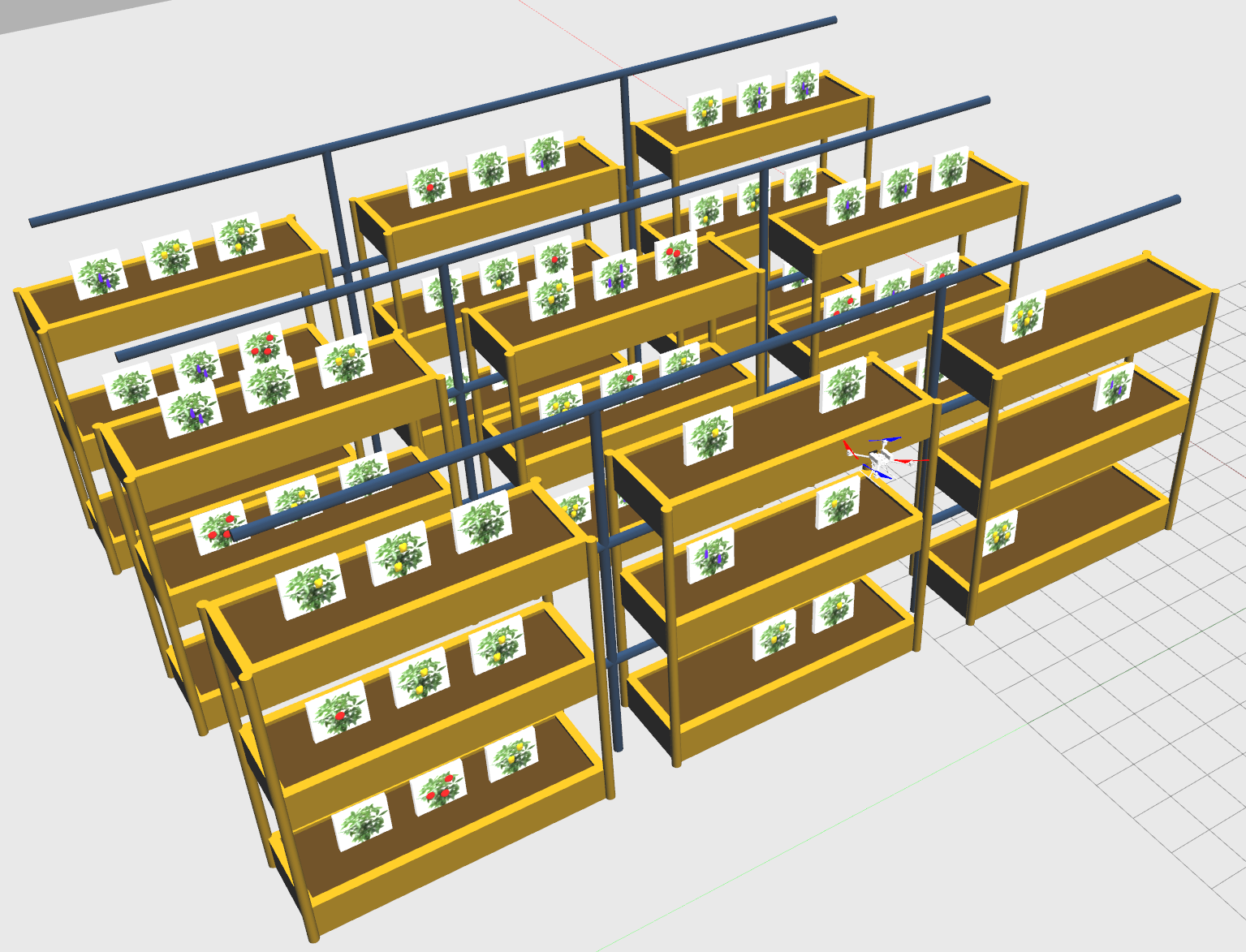}
    \caption{Example of scene configuration in Gazebo simulation.}
    \label{fig:sim-world}
\end{figure}



In our work, we are using a Gazebo \cite{Gazebo} simulation environment prepared by the organisers of the UAV competition. 
It simulates a warehouse with 27 shelves containing ''bushes'' of one of three types of plants: tomato (red fruit), pepper (yellow fruit) and eggplant (purple fruit).
Each plant can contain a certain number of fruits, but it is also acceptable to have no fruits.
An example of the configuration of the simulation environment is shown in Figure \ref{fig:sim-world}.
The objective of a single mission is to count the sum of all fruits of a given type from the given locations.
To do this, a flight route must be planned that minimises the distance travelled and the time taken to complete the task, taking into account the need to inspect both sides of each plant bed under investigation.

\begin{figure}
    \centering
    \includegraphics[width=1\linewidth]{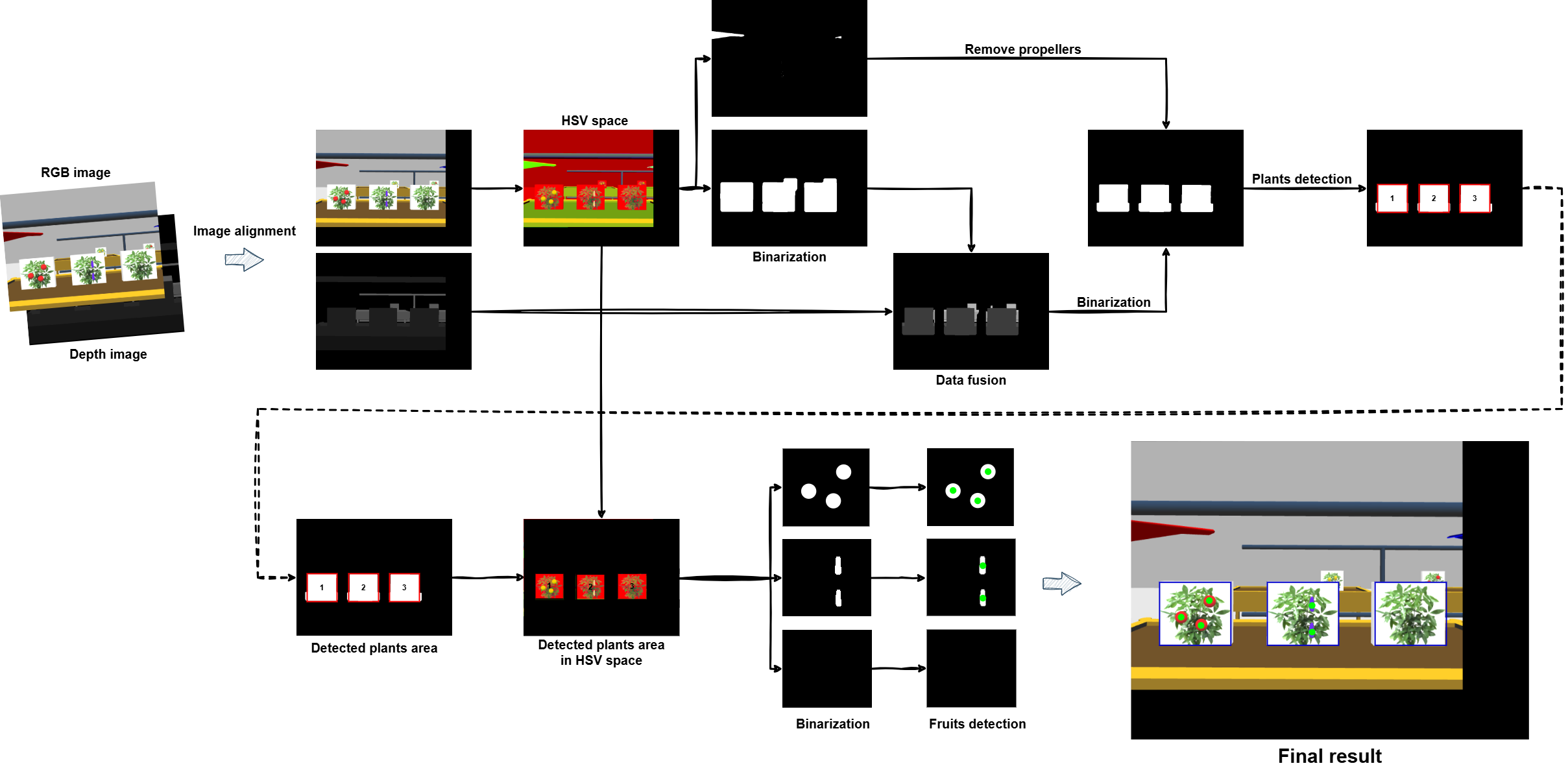}
    \caption{Scheme of our algorithm for fruits detection in simulation environment.}
    \label{fig:alg-sim}
\end{figure}

The basic problem in this situation is to detect and locate fruit of a particular type.
Our solution consists of a series of classical image processing operations, summarised in Figure \ref{fig:alg-sim}.


The starting point is the image from the RGB camera and the depth sensor.
In a first step, we convert the RGB image to HSV (Hue, Saturation, Value) space, in which it is easier to detect the fruits of interest (tomato, pepper and eggplant), as well as the green parts of the plant and the white background of the board.
Next, we detect the areas where the plants are located (plant beds).
To do this, we combine the masks of the white and green parts and subject them to a morphological dilation operation with a window size of $1 \times 25$ and $ 25 \times 1$ and median filtering. 
In the next step, we use the resulting mask to cut out areas from the depth map where plant beds may be found.
We segment the plant bed candidates and analyse them by size. 
If the area exceeds a preset threshold (100,000 pixels), we consider the area valid and save it for later fruit detection.
The use of a depth map makes it easy to avoid errors caused by false detection of fruit visible in the background (in successive rows of plant beds).
In parallel, the HSV image is analysed for the visibility of drone propellers, which can also affect the correctness of the image analysis and fruit detection.
Finally, a binary fusion is performed.

The analysis of individual plant beds is implemented as follows.
We cut out a fragment of the image in HSV space corresponding to a given location (left side, centre and right side in the camera frame). 
On the obtained fragment, we perform segmentation based on empirically selected thresholds.
We refine the obtained mask with a set of filters: median, erosion and dilation, all with a $3 \times 3$ kernel.
On the resulting image, we finally perform object analysis -- fruit counting.
For each of the detected connected components, we determine the coordinates of the bounding box, its area and the area of the object.
For objects satisfying the conditions for the minimum field and size of the bounding box (empirically 100 and 200 pixels, respectively), we determine the distance transform, whose result we binarise with a threshold of 0.7 and re-analyse.
If, in this case, the field size is greater than 10 pixels, we consider the object to be a valid fruit.
The distance transform solution used makes it possible to detect objects that are ‘compact’ and, at the same time, to separate fruits that are in contact with each other (occurring in some situations).
After the described transform binarisation procedure, only the central fragments of the counted fruits are detected.

The final component of the counting algorithm is to combine the results for the same plant bed seen from two different sides.
In particular, according to the conditions of the competition, some fruits may only be visible from one side and others from both.
Due to the high precision of the detection and location of the objects, the merging is relatively simple and requires basic geometrical operations and checking if any fruit is detected ‘on the other side’.

Another problem to be solved in this scenario is the correct planning of the flight trajectory.
First, we determine all 3D points and the corresponding spatial orientations of the drone that must be passed during the mission, based on the given plant locations.
This gives us a well-defined travelling salesman problem, which we solve using the Fast-TSP algorithm \cite{fast-tsp}.
We take into account the different costs of flying between points, including the need to avoid beds.
From the resulting order of points to visit, we prepare a reference trajectory, which is executed by the TOPP-RA tracker \cite{toppra}.
We also try to minimise the distortion of the recorded camera video stream caused by the movement of the drone (e.g. blurring, altered perspective, image rotation) near the desired locations.
To this end, when we prepare a reference trajectory, we interpolate between given points using Chebyshev nodes.



\section{Results and discussion}\label{sec:results}


To test and evaluate our solution, we used the simulation environment described above to create a test setup. 
We prepared it fully automatically with randomly selected and configured plant beds.
In it, we ran various flights according to the specially developed scheduler.
A single mission is defined by the name of the plant and the order of the shelves to be checked, e.g. \texttt{Tomato 2 3 8 14 25}.
Other factors such as fruit size and lighting conditions were the same for all missions.
However, this does not mean that all the conditions for detecting a particular plant (e.g. camera distance from the plant) were the same.
This depends on the trajectory planning and execution, which can vary depending on the mission specification.

With this in mind, the scheduler determined the following parameters before starting each test:
\begin{enumerate}
    \item Type of plant whose fruit is to be counted (one of the following: tomato, pepper, eggplant).
    \item Number of beds to be checked during the mission (from 1 to 27).
    \item Bed indexes to be checked during the mission (ascending sequence of values between 1 and 27).
\end{enumerate}
In total, we have $3 \times 2^{27}$ different missions.
Due to the large number of possibilities, we decided to run the tests only on a limited set of 500 missions randomly generated by the scheduler.
The mission specification presented is the same as that proposed by the organisers of the UAV Competition.
The only difference is the number of test cases.
During the competition, the final score was determined by the average of the best 4 out of 6 missions.
Evaluating more cases allowed us to potentially find some bugs that could only manifest themselves in a few specific cases.


In evaluating our fruit counting system, we followed the approach suggested by the organisers of the UAV Competition.
We describe the metrics we used in Subsection \ref{subsec:eval metrics}.
We present the results obtained with them, together with a discussion, in Subsection \ref{subsec:eval results}. 

\subsection{Evaluation metrics}\label{subsec:eval metrics}


The total score for the mission accomplished is the sum of the four components according to the equation below:
\begin{equation}
    \label{eq:eval points}
    p = p_f + p_t + p_d - p_c
\end{equation}
where:
\begin{itemize}
    \item $p_f = 50(1 - 4\frac{|c_r - c_t|}{c_t})$ -- points for reporting ($c_r$) the correct number of fruits ($c_t$),
    \item $p_t = 25e^{(1-\frac{t_m}{t_b})}$ -- points for mission time ($t_m$) in relation to the base time ($t_b$),
    \item $p_d = 25e^{2(1 - \frac{d_m}{d_b})}$ -- points for measured distance covered during mission ($d_m$) in relation to the base distance ($d_b$),
    \item $p_c = 25k$ -- penalty points for total number ($k$) of collisions with elements of the environment and fllying off the allowed space.
\end{itemize}
The metric presented contains some reference values $t_b$ and $d_b$, which are the same for all missions performed.
The actual flight time and the distance covered are compared to these values.
The evaluation $t_m = t_d$ or $d_m = d_b$ results in 25 points for the component.
The reference values can be interpreted as the desired flight time and distance travelled for a single mission.
In general, they will obviously vary depending on the configuration of a given mission, in particular the number of locations to be visited.
Nevertheless, in order to be consistent with the organisers of the UAV Competition, we have taken the reference time and distance parameters in the metric \eqref{eq:eval points} as constants $t_b = 100 [s]$ and $d_b = 150 [m]$.
These can then be interpreted as averages of the desired flight time and distance travelled over a limited set of all possible missions.
This approach allows us to relate the obtained results to some extent to other contestants.

\subsection{Evaluation results}\label{subsec:eval results}

\begin{figure}
    \centering
    \subfloat[Bed distribution for all test missions.]{\label{fig:beds distribution}
    \centering
    \includegraphics[width=0.48\linewidth]{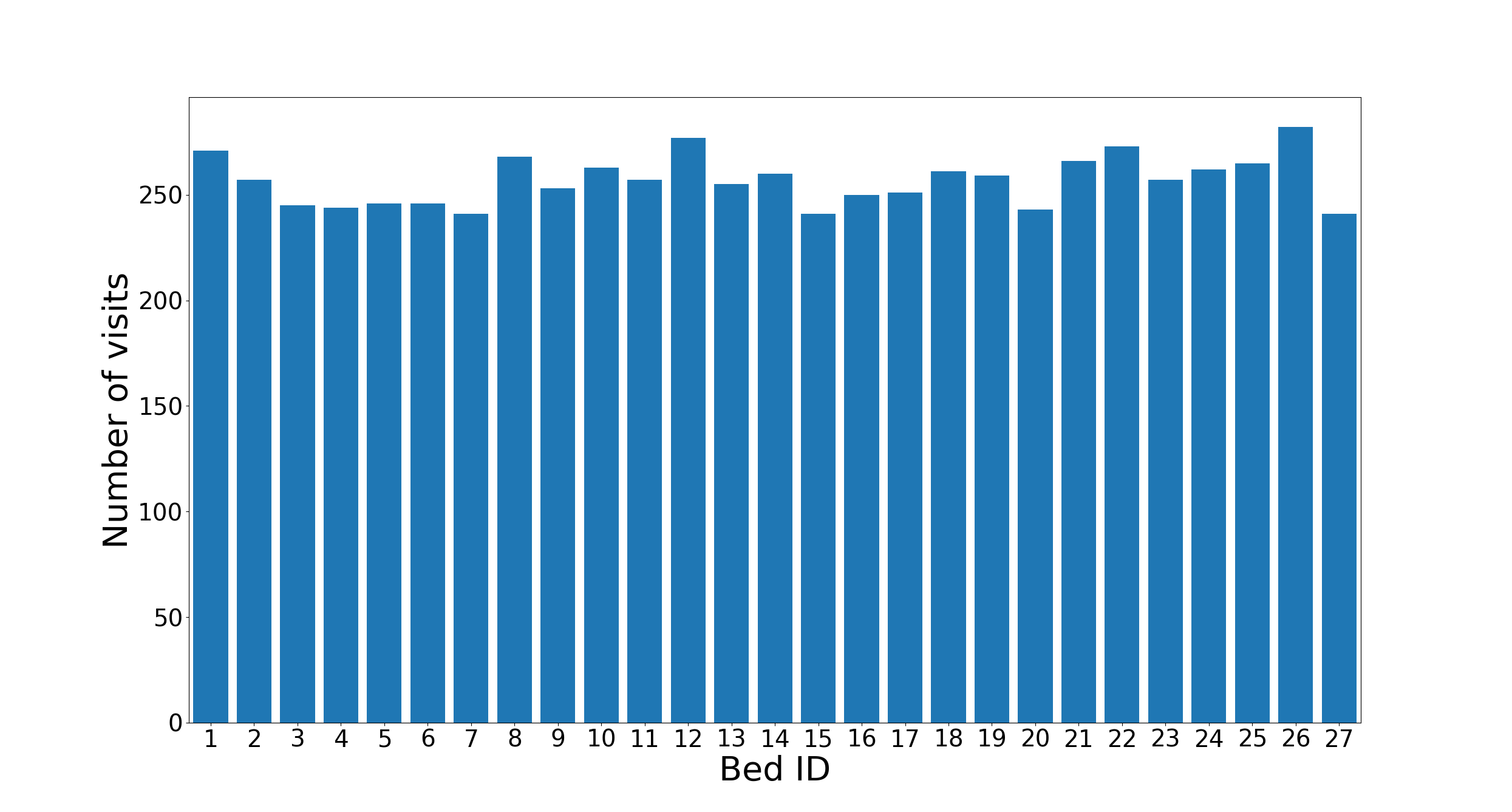}
    }
    \hfill
    \subfloat[Mission's length distribution for all test missions.]{\label{fig:length distribution}
    \centering
    \includegraphics[width=0.48\linewidth]{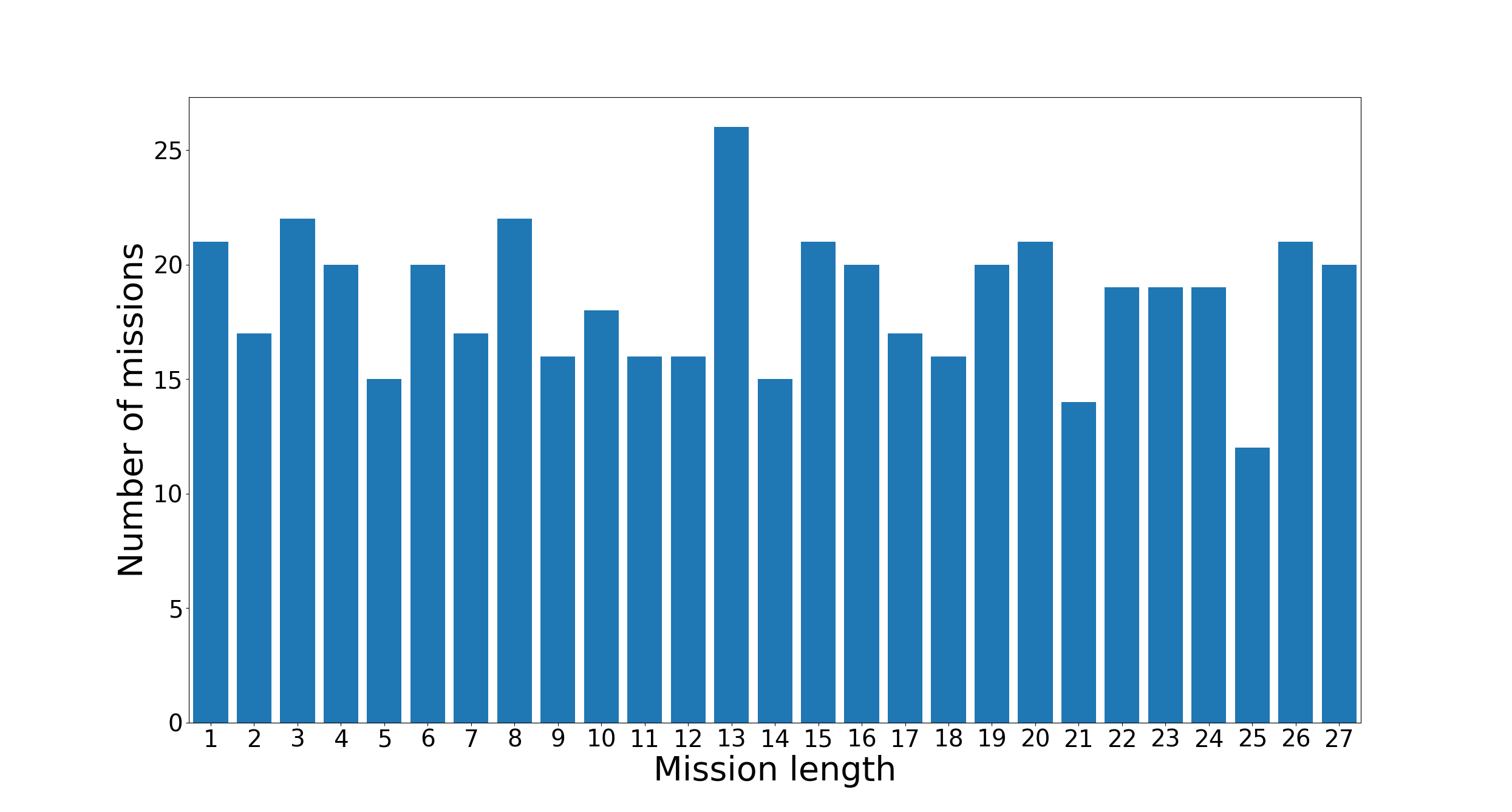}
    }
    \hfill
    \subfloat[Fruit's type distribution for all test missions.]{\label{fig:fruit distribution}
    \centering
    \includegraphics[width=0.68\linewidth]{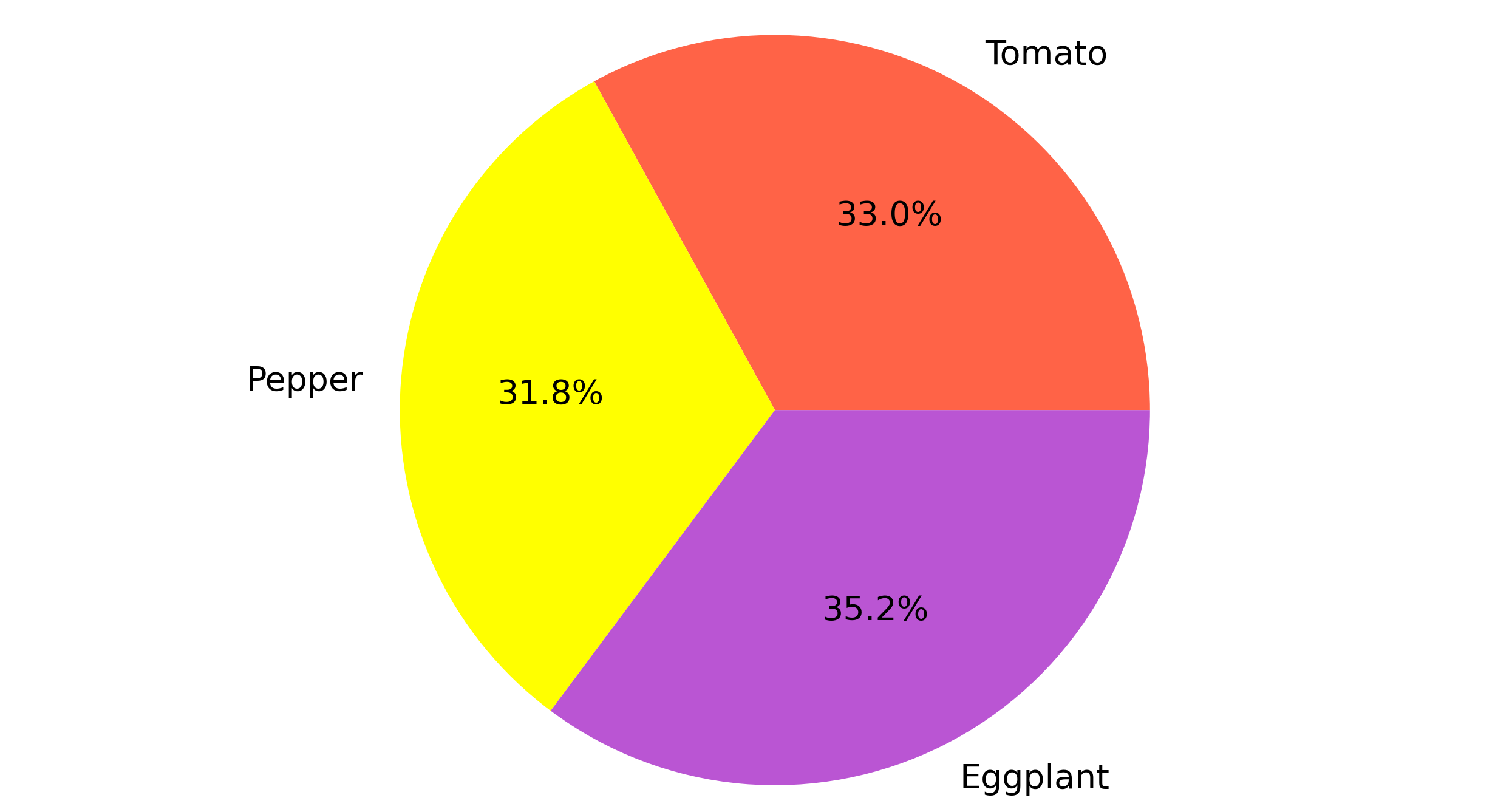}
    }
    \caption{Test missions parameters distribution for 500 total number of cases.}
    \label{fig:tests statistics}
\end{figure}


We evaluated our solution on a set of 500 missions.
They were randomly selected and we show the distribution of their basic parameters in Figure \ref{fig:tests statistics}.
Both the beds visited (Figure \ref{fig:beds distribution}) and the types of plants counted (Figure \ref{fig:fruit distribution}) were almost uniformly covered.
Only the length distribution (Figure \ref{fig:length distribution}) shows a slightly greater variation in frequency, but all possible cases were sufficiently covered.
In view of this, it must be concluded that the test cases generated by us are authoritative and reflect the entire set of missions.

\begin{table}[]
    \centering
    \caption{Average evaluation results for total number of 500 missions. All symbols according to \eqref{eq:eval points}.}
    \begin{tabular}{|c|c|c|c|c|}
        \hline
        $\bar{p_f}$ & $\bar{p_t}$ & $\bar{p_d}$ & $\bar{p_c}$ & $\boldsymbol{\bar{p}}$ \\
        \hline
        49.88 & 13.51 & 23.88 & 0.00 & \textbf{87.27} \\
        \hline
    \end{tabular}
    \label{tab:eval results}
\end{table}

The results of our evaluation are shown in the Table \ref{tab:eval results}.
The first thing to note is the complete absence of a collision penalty ($\bar{p_c}=0.00$).
Our solution therefore shows a high level of safety, which is also a fundamental criterion for most UAV missions.
The average fruit count score ($\bar{p_f}=49.88/50.00$) is also close to the maximum.
This demonstrates the high performance of the vision-based detection and localisation algorithm we developed.
For the analysis of the other two components, i.e. points for flight time and distance covered, the form of the metric \eqref{eq:eval points} and the comments we made in Subsection \eqref{subsec:eval metrics} should be kept in mind.
Nevertheless, the average points per distance covered is close to the reference value ($\bar{p_d}=23.88/25.00$).
This indicates a very good performance of our trajectory planner.
At the same time, the average score for flight time is significantly below the reference value ($\bar{p_t}=13.51/25.00$).
This could be an indication of a mission execution speed that is too low, or an increased loss of time during taking pictures of fruits.
Indeed, in both cases we decided to take a conservative approach to ensure collision-free flights and correct fruit counts.
Our decisions can be justified by the results shown in the Table \ref{tab:eval errors}.
The system that we developed is highly resilient to potential problems associated with missing required plant beds, collisions or counting errors.
It is worth noting that our system uses only classical algorithms for both image recognition and UAV control.
This makes it well suited for real-time video processing with an on-board embedded computing platform.
Compared to other solutions, which are mainly based on AI models, our system has lower computational and power requirements, which is crucial for UAV-related applications.
It is therefore fully explainable with predictable and consistent behaviour, a key factor in mission safety.

\begin{table}[]
    \centering
    \caption{Error sources for total number of 500 missions.}
    \begin{tabular}{|l|c|}
        \hline
        Missions with missing beds & 0 (0.00 \%) \\
        \hline
        Missions with collisions & 0 (0.00 \%) \\
        \hline
        Missions with counting errors & 2 (0.40 \%) \\
        \hline
    \end{tabular}
    \label{tab:eval errors}
\end{table}


As the AVADERAGH team, we submitted our solution to the UAV Competition accompanying the ICUAS 2024 conference.
In the official evaluation by the organisers, we achieved an average of 84.83 points with the best of 4 out of 6 missions evaluation\footnote{\url{https://github.com/larics/icuas24_competition/discussions/52}}.
This placed us 6th in a field of 23 teams and enabled us to progress to the final of the competition, where we achieved the 4th place.
However, the detailed specification of the final round is slightly different from the simulation phase.
It focuses on the recordings registered by the organisers.
Therefore, the control part is omitted in this case.
In addition, the UAV is equipped with a different set of sensors.
Most importantly, the depth camera is missing and a LiDAR sensor is used instead.
Therefore, the final stage is beyond the scope of this paper.

This result demonstrates the competitiveness of our solution.
It also shows that our evaluation and its results, based on a much larger test set (500 vs. 6 missions), are in line with the organisers' evaluation. 

\section{Summary}\label{sec:summary}


In this paper we presented a fruit counting system using UAVs.
It meets the requirements of the UAV competition organised as part of the ICUAS 2024 conference.
Fruits are detected using a vision algorithm developed by us, which operates on video streams from a classical RGB camera and a depth sensor.
It uses classical image processing algorithms in the HSV colour space.
Our system also allows trajectory planning using the Fast-TSP algorithm and trajectory execution by the TOPP-RA tracker.


We tested our system in a simulation environment provided by the organisers of the UAV Competition on a randomly generated set of 500 missions.
We scored them using our own algorithm, prepared on the basis of the UAV Competition organisers' guidelines.
We obtained an average score of 87.27/100 points.
In addition, our system demonstrated high resistance to potential hazards such as omitting a given location, collisions with the environment or incorrect fruit counts.
We submitted our solution to the UAV Competition, where we placed 6th with a score of 84.83/100 points and advanced to the finals.


The results show that our proposed approach is highly competitive compared to other solutions.
We therefore plan to develop it further by carrying out real-life tests.
We envisage starting the trials in a suitably prepared environment, with the fruit represented by artificial elements such as balls of a particular colour, and then gradually increasing the level of difficulty.
We also plan to implement the whole system on a heterogeneous embedded computing platform, e.g. SoC FPGA or Nvidia Jetson, to enable autonomous mission execution by the UAV.

\bibliography{hsz_mw_rm_tk_automation_2025}

\end{document}